# IMPLEMENTATION OF TORQUE CONTROLLER FOR BRUSHLESS MOTORS ON THE OMNI-DIRECTIONAL WHEELED MOBILE ROBOT

*Piyamate Wasuntapichaikul[1], Kanjanapan Sukvichai[2] and Yodyium Tipsuwan[1]*

[1]Dept. of Computer Engineering, Kasetsart University, Thailand
[2]Dept. of Electrical Engineering, Kasetsart University, Thailand

## ABSTRACT

The major issue for the wheeled mobile robot is the low level controller gains tuning up especially in the robot competition. The floor surface can be damaged by the robot wheels during the competition, therefore the surface coefficient can be changed over time. PI gains have to be tuned before every match along the competition. In this research, the torque controller is defined and implemented in order to solve this problem. Torque controller consists of a PI controller for the robot wheel's angular velocity and a dynamic equation of brushless motor. The motor dynamics can be derived from the energy conservation law. Three different carpets, which have the different friction coefficients, are used in the experiments. The robot wheel's angular velocity profiles are generated from the robot kinematics with different initial conditions. The output paths of the robot with the torque controller are compared with the output paths of the robot with regular PI controller when the same wheel angular velocity profiles are applied. The results show that the torque controller can provide a better robot path than the normal PI controller.

## 1. INTRODUCTION

Omni-directional wheel robot is one of the most popular mobile robots which is used in many research fields because of its maneuverability. Omni-directional wheel allows robot to move along different paths with efficiency. Robot with four motorized wheels is generally used by many researchers and also in the small size league in RoboCup competition because they represent a good real life example of the multi-agent robot cooperation. Skuba, which is a robot team from Kasetsart University [1], joined the small size league since 2005 and got the champion from RoboCup 2009. The major problem for robot teams is the manual calibration of the low level controller gains. The surface parameters are changed over time because the carpet is damaged from robot wheels. Therefore, all the low level controller gains have to be adjusted every match. Torque control scheme is implemented in this research in order to solve this problem. Torque controller consists of a PI controller and a torque converter. Maxon brushless motor model is derived and used in the control law. The difference between each wheel's angular velocity and the desired angular velocity is fed as input to the torque controller and the output is sent as the control signal to each motor. The experiments are setup in order to show the advantages of the torque controller to the regular PI controller for controlling the angular velocity of the motor.

## 2. ROBOT DYNAMICS

The dynamics of a robot is derived in order to provide information about its behavior. Kinematics alone is not enough to see the effect of inputs to the outputs because the robot kinematics lacks information about robot masses and inertias. The dynamic of a robot can be derived by many different methods such as Newton's law [2] [3] and Lagrange equation [4]. In this paper, Newton's law is used to solve the dynamic equation of the robot. The interested mobile robot consists of four omni-directional wheels as shown in Figure 1. Newton's second law is applied to robot chassis and the dynamic equation can be obtained as (1) though (3).

$$\ddot{x} = \frac{1}{M}(-f_1 \sin\alpha_1 - f_2 \sin\alpha_2 + f_3 \sin\alpha_3 + f_4 \sin\alpha_4) - \bar{f}_f\big|_x \quad (1)$$

$$\ddot{y} = \frac{1}{M}(f_1 \cos\alpha_1 - f_2 \cos\alpha_2 - f_3 \cos\alpha_3 + f_4 \cos\alpha_4) - \bar{f}_f\big|_y \quad (2)$$

$$J\ddot{\theta} = d(f_1 + f_2 + f_3 + f_4) - T_{trac} \quad (3)$$

where,

$\ddot{x}$ is the linear acceleration of the robot along the x-axis of the global reference frame

$\ddot{y}$ is the linear acceleration of the robot along the y-axis of the global reference frame

$M$ is the total mass of the robot

$f_i$ is the motorized force of the wheel i

$\bar{f}_f$ is the friction force vector

$\alpha_i$ is the angle between wheel i and the robot x-axis

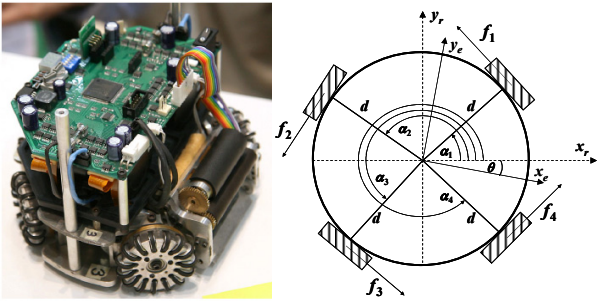

Fig. 1. The robot structure.

$\ddot{\theta}$ is the angular acceleration of the robot about the z-axis of the global reference frame
$J$ is the inertia of the robot
$d$ is the distance between wheels and the robot center
$T_{trac}$ is the traction torque of the robot

The robot inertia, friction force and traction torque are not directly found from the robot mechanical configuration. These parameters can be found by experiments. The robot inertia is constant for all different floor surfaces while the friction force and traction torque are changed according to floor surfaces. The friction force and traction torque are not necessarily found at this point because these two constraints are different for each floor surface and their effect can be reduced by using the control scheme which is discussed in the next topic. The motorized forces are driven by brushless motors with $\tau_{m1}$, $\tau_{m2}$, $\tau_{m3}$, and $\tau_{m4}$, therefore the dynamic equation of the robot becomes

$$\ddot{x} = \frac{1}{M}(-\frac{\tau_{m1}}{r}\sin\alpha_1 - \frac{\tau_{m2}}{r}\sin\alpha_2 + \frac{\tau_{m3}}{r}\sin\alpha_3 + \frac{\tau_{m4}}{r}\sin\alpha_4) - \bar{f}_f\big|_x \quad (4)$$

$$\ddot{y} = \frac{1}{M}(\frac{\tau_{m1}}{r}\cos\alpha_1 - \frac{\tau_{m2}}{r}\cos\alpha_2 - \frac{\tau_{m3}}{r}\cos\alpha_3 + \frac{\tau_{m4}}{r}\cos\alpha_4) - \bar{f}_f\big|_y \quad (5)$$

$$\ddot{\theta} = \frac{d}{J}(\frac{\tau_{m1}}{r} + \frac{\tau_{m2}}{r} + \frac{\tau_{m3}}{r} + \frac{\tau_{m4}}{r}) - T_{trac} \quad (6)$$

where,
$r$ is a wheel radius

Equation (4) though (6) show that the dynamics of the robot can be directly controlled by using the motor torques.

## 3. MOTOR MODEL AND TORQUE CONTROLLER

A Maxon brushless motor is selected for the robot. The dynamic model of the motor can be derived by using the energy conservation law as shown in [5]. The dynamic equation for the brushless motor is

$$u \cdot \frac{\tau_m}{k_m} = \frac{\pi}{30,000} \cdot \dot{\phi} \cdot \tau_m + R \cdot \left(\frac{\tau_m}{k_m}\right)^2 \quad (7)$$

where,

$u$ is the input voltage
$\tau_m$ is the output torque of the motor
$k_m$ is the torque constant of the motor
$\dot{\phi}$ is the angular velocity of the motor
$R$ is the coil resistance of the motor

Equation (7) shows that input voltage has a linear relationship with the output torque at specific angular velocity. This equation is needed to be modified to a simple version by using parameters relationship of the Maxon motor. The final dynamic equation of the motor is

$$\tau_m = \left(\frac{k_m}{R}\right) \cdot u - \left(\frac{k_m}{R \cdot k_n}\right) \cdot \dot{\phi} \quad (8)$$

where,
$k_n$ is the motor speed constant

Equation (8) shows the direct relation of the control signal $u$ and the output torque $\tau_m$ at the specific angular velocity $\dot{\phi}$. The control scheme is set using the discrete Proportional-Integral control law and torque dynamic equation (8). From the experiment, the measured angular velocity has a high frequency noise, therefore the low pass FIR filter is required. The error between desired angular velocity and real filtered angular velocity of each wheel is the input of the PI controller with the PI gains $kp$ and $ki$ respectively. The controller is shown in Figure 2 and the control law can be described as (9) through (11).

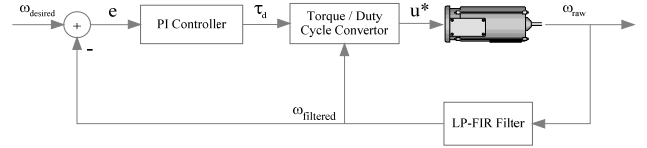

Fig. 2. The torque controller.

$$err[j] = \dot{\phi}_{desired}[j] - filtered[\dot{\phi}_{real}[j]] \quad (9)$$

$$\tau_d[j] = k_p \cdot err[j] + k_I \cdot \sum_{j=1}^{N}(err[j]) \quad (10)$$

$$u^*[j] = \frac{\tau_d[j]}{\left(\frac{k_m}{R}\right) \cdot V_{cc} - \left(\frac{k_m}{R \cdot k_n}\right) \cdot filtered[\dot{\phi}_{real}[j]]} \quad (11)$$

where,
$N$ is the number of samples
$V_{cc}$ is the power supply voltage of the motor driver
$u^*$ is the duty cycle of the PWM control signal

Since the control signal for the motor is Pulse Width Modulated (PWM) signal, the output of the controller has to be the duty cycle for PWM signal generator. Equation (11) shows that the duty cycle of the control signal is the ration of the desired torque divided by the maximum torque of the motor at the particular angular velocity.

## 4. EXPERIMENT

To achieve the purpose of this research, the PI torque controller is implemented and applied to the wheeled mobile robot. Skuba robot is selected for all experiments. The robot consists of four omni-directional wheels which are driven by 30 watt Maxon flat brushless motors. Each motor is equipped with a 360 CPR optical encoder to provide signals for speed measurement. All the robot hardware are controlled by a single-chip Spartan-3 FPGA from Xilinx. The FPGA contains a soft 32-bit microprocessor core that runs at 30 MIPS and interconnected peripherals. This embedded processor executes the low level motor control loop with 600Hz control loop cycle. The interested output of this research is the error of robot position profile when the same input velocity is applied for the different surfaces. Three different carpets are used in the experiment. The normal angular velocity control using PI controller and the torque control using PI controller are applied to the selected robot. Both controller gains are tuned until the robot can perform the same behavior on one of the test surface. This controller gains are used for all experiments. By changing the trajectory profile of the robot, the floor surfaces, the advantages of each controller are revealed. Robot and motor parameters are shown in Table 1.

| $M$ | 1.5 kg |
|---|---|
| $J$ | 0.0192 kg/m$^2$ |
| $d$ | 78.95 mm |
| $[\alpha_1, \alpha_2, \alpha_3, \alpha_4]$ | [33,147,225,315] degree |
| $r$ | 25.4 mm |
| $V_{cc}$ | 14.8 V |
| $k_m / R$ | 0.02125 $Nm/A\cdot\Omega$ |
| $k_m / (R\cdot k_n)$ | 0.0005426 $Nm\cdot V\cdot s/A\cdot\Omega\cdot rad$ |

Table 1. Robot parameters.

| Case | $\theta(0)$ | $\dot{x}$ | $\ddot{x}$ |
|---|---|---|---|
| 1 | 90 | 2 | 3 |
| 2 | 90 | 1.5 | 2 |
| 3 | 45 | 0.8 | 1 |
| 4 | 0 | 0.8 | 1 |

Table 2. Four set of parameters tested in each surface.

Trapezoidal trajectories for robot are generated by using the robot kinematic equation with different conditions. The input angular velocity profile is shown in Figure 3.

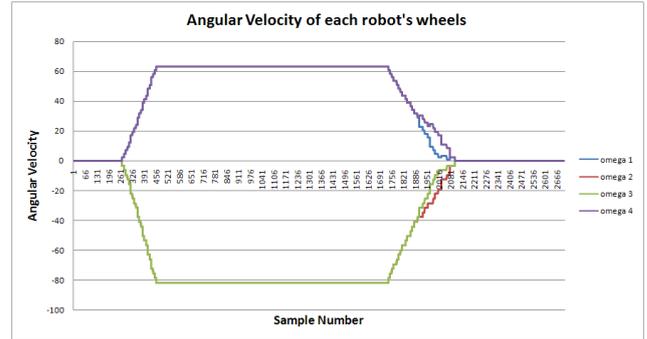
Fig. 3. Input velocity profiles for each motor.

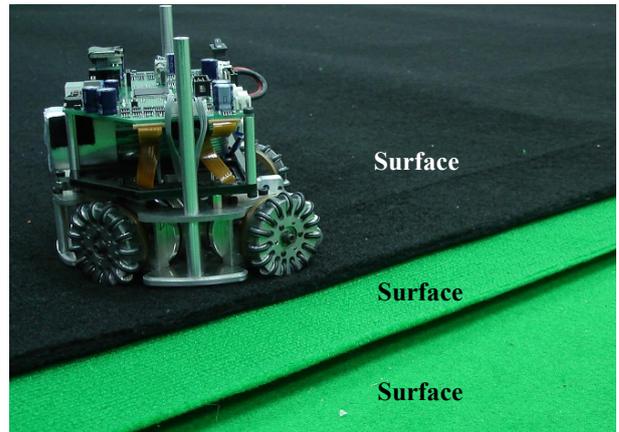
Fig. 4. Three selected carpets.

The examples of conditions for each velocity profile are shown in Table 2, where $\dot{x}$ and $\ddot{x}$ are desired velocity of the robot along the robot's x-axis. $\theta(0)$ is the initial orientation of the robot. The desired velocity profiles are directly applied to the robot on three different surfaces without any global vision feedback controller. Figure 4 shows the selected carpets which are used in this experiment. The carpet surfaces have a different friction coefficient ($\mu$) with $\mu_1 < \mu_2 < \mu_3$. The angular velocity of the robot is recorded by FPGA board while the position is recorded through the bird eye view video camcorder.

The angular velocity output of motors are collected and compared with the output from same motors but different carpets. Figure 5 shows the angular velocity output of a motor which is controlled by the torque controller while Figure 6 shows the output angular velocity of a motor controlled by regular PI controller. From the experiment, a motor with torque controller can maintain its velocity when the surface frictions are changed. For a motor with normal PI controller, the output is swing dramatically when it runs on different surfaces. The result shows that a motor with torque controller has a better tracking response than a motor with regular PI controller especially when the motor is braking.

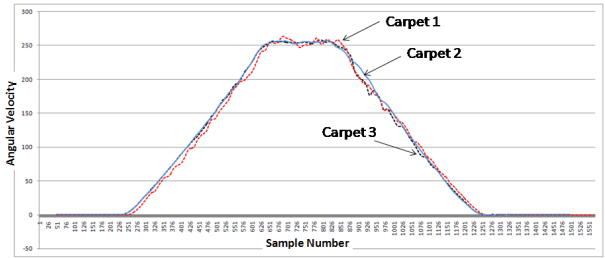

Fig. 5. Angular velocity of a motor with torque controller.

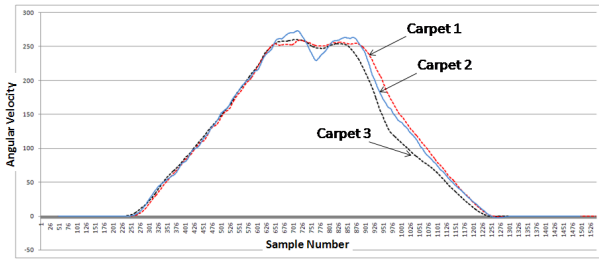

Fig. 6. Angular velocity of a motor with regular PI controller.

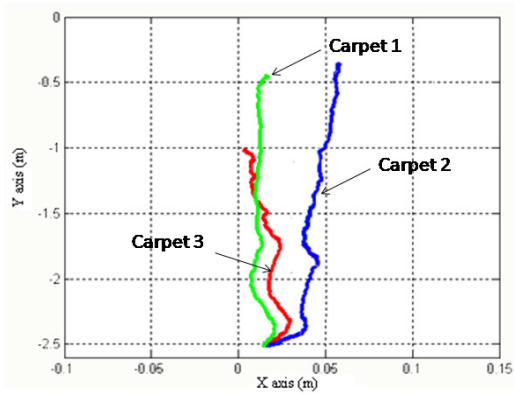

Fig. 7. Robot path when using torque controller.

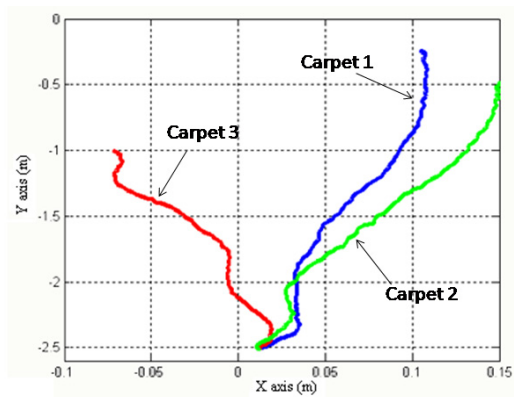

Fig. 8. Robot path when using regular PI controller.

Figure 7 and 8 show the trajectory of the robot on different carpets. Robot with the torque controller has a better performance when compared with a robot with the regular PI controller. Moreover, the robot path when controlled by the torque controller on the first carpet is close to the second and third carpet while the robot path when controlled by the normal PI controller on the first carpet is totally different from the second and third carpet. This result yields that if the robot is tuned up for one surface friction with the torque controller, it is possible to have a very close result when it runs on different surfaces.

## 5. CONCLUSION

The dynamic equation of the mobile robot reveals the relationship of the robot acceleration and input motorized torques. Due to this fact, the dynamic equation of the motor torque is derived and used in order to construct the torque controller. By adding the torque converter block to the regular PI controller for the wheel angular velocity control law, the torque controller is defined. The experiments are set up in order to compare the robot behavior when using the torque controller and normal PI controller. The result shows the output paths of the robot when the surface friction coefficients are changed. The position errors of the robot on different surfaces when controlled by the torque controller are always smaller than the robot controlled by the regular PI controller. Therefore, the torque controller is more convenient to use in the small size league robot competition for the low level motor controller because the controller can provide a better path when the surface coefficient is changed during the competition and it can reduce the robot tuning time before matches.